\documentclass[11pt,a4paper]{article}
\usepackage[hyperref]{emnlp2018}
\usepackage{times}
\usepackage{latexsym}
\usepackage{booktabs}

\usepackage{url}

\usepackage{array}
\newcolumntype{K}[1]{>{\centering\arraybackslach}p{#1}}

\aclfinalcopy 

\title{Teaching Syntax by Adversarial Distraction}

\author{Juho Kim \\
  University of Illinois \\
  Urbana, IL \\
  {\tt juhokim2@illinois.edu} \\\And
  Christopher Malon \\
  NEC Laboratories America \\
  Princeton, NJ \\
  {\tt malon@nec-labs.com} \\ \And
  Asim Kadav \\
  NEC Laboratories America \\
  Princeton, NJ \\
  {\tt asim@nec-labs.com} \\}

\date{}

\begin{document}
\maketitle
\begin{abstract}
Existing entailment datasets mainly pose problems which can be
answered without attention to grammar or word order.
Learning syntax requires comparing examples where different
grammar and word order change the desired classification.
We introduce several datasets based on synthetic transformations of
natural entailment examples in SNLI or FEVER, to teach
aspects of grammar and word order.
We show that without retraining, popular entailment models are unaware
that these syntactic differences change meaning.
With retraining, some but not all popular entailment models can
learn to compare the syntax properly.
\end{abstract}

\section{Introduction}

Natural language inference (NLI) is a task to identify the entailment
relationship between a premise sentence and a hypothesis sentence.
Given the premise, a hypothesis may be true (entailment),
false (contradiction), or not clearly determined (neutral). NLI is
an essential aspect of natural language understanding. The release of
datasets with hundreds of thousands of example pairs, such as SNLI
\citep{SNLI15} and MultiNLI \citep{multinli18}, has enabled the development
of models based on deep neural networks that have achieved near human level
performance.

However, high accuracies on these datasets do not mean that the NLI
problem is solved.  Annotation artifacts
make it possible to correctly guess the label for many hypotheses
without even considering the premise
\citep{annotation_artifacts18, hypothesisOnly18, hiddenbias18}.
Successful trained systems can be disturbed by
small changes to the input \citep{breakingNLI18, stresstest18}. 

In this paper, we show that existing trained NLI systems are mostly unaware
of the relation between syntax and semantics, particularly of how word
order affects meaning.  We develop a technique, ``adversarial
distraction,'' to teach networks to properly use this information.
The adversarial distraction technique consists of creating pairs of
examples where information matching the premise is present in the hypothesis
in both cases, but differing syntactic structure leads to different
entailment labels.
We generate adversarial distractions automatically from SNLI and an NLI
dataset derived from FEVER \citep{fever18}, thus augmenting the datasets.
We observe the behavior of several existing NLI models on the
added examples, finding that they are mostly unaware that the syntactic
changes have affected the meaning.
We then retrain the models with the added examples, and find whether these
weaknesses are limitations of the models or simply due to the lack of
appropriate training data.

\section{Related work}


{\bf Datasets for NLI}:
SNLI and MultiNLI are both based on crowdsourced annotation.  In SNLI
all the premises came from image captions
\citep{imageCaption14}, whereas MultiNLI collected premises from
several genres including fiction, letters, telephone speech, and a
government report.  SciTail \cite{scitail18} constructed more complicated
hypotheses based on multiple-choice science exams, whose premises were taken
from web text.  More recently, FEVER introduced a fact
verification task, where claims are to be verified using all of Wikipedia.
As FEVER established ground truth evidence for or against each claim,
premises can be collected with a retrieval module and
labeled as supporting, contradictory, or neutral for an NLI dataset.


{\bf Neural network based NLI systems}: Dozens of neural network based
models have been submitted to the SNLI leaderboard.  
Some systems have been developed based on sentence representations
\citep{infersent17, SenEncMultiNLI17}, but most common models apply
attention between tokens in the premise and hypothesis.  We focus on
three influential models of this kind: Decomposable Attention
\citep{DecAtt16}, ESIM \citep{esim17}, and a pre-trained transformer
network \citep{openai_trans18} which obtains state-of-the-art
results for various NLI datasets including SNLI and SciTail.

{\bf Adversarial examples for NLI systems}:
\citet{jiaLiang17} introduced the notion of distraction for
reading comprehension systems by trying to fool systems for SQuAD \citep{squad}
with information nearly matching the question, added to the end of
a supporting passage. \citet{breakingNLI18} showed that 
many NLI systems were confused by hypotheses that were identical to
the premise except for the replacement of a word by a synonym, hypernym,
co-hyponym, or antonym. \citet{stresstest18} found that
adding the same strings of words to NLI examples without changing the logical
relation could significantly change results, because of word overlap,
negation, or length mismatches.

Other work \citep{adventure18, GenAdversarial18} aimed to improve
model robustness in the framework
of generative adversarial networks \citep{GAN14}. \citet{debuggingNLP18} 
generated semantically equivalent examples using
a set of paraphrase rules derived from a machine translation model.
In contrast to these kinds of adversarial examples, we focus on the model not
being sensitive enough to small changes that do change meaning.

\section{Teaching Syntax}

Our adversarial examples attack NLI systems from a new direction: not in
their failure to capture the relations between words as in \citet{breakingNLI18}, 
but their failure to consider the syntax and
word order in premises and hypotheses to decide upon the entailment relation.
As position-agnostic approaches such as Decomposable Attention and
SWEM \citep{swem} provide competitive baselines to existing datasets,
the power of models to interpret token position information has not been
rigorously tested.

\begin{table*}[t!]
\begin{center}
\begin{tabular}{lcccccc}
\hline 
 & \multicolumn{3}{c}{SNLI} & \multicolumn{3}{c}{FEVER} \\ 
 & original & passive & passive rev & original & person rev & birthday \\
\hline \hline
\# train      & 549,367 & 129,832 & 39,482 & 602,240 & 3,154 & 143,053 \\
\# validation &   9,842 &   2,371 &    724 &  42,541 &    95 &   9,764 \\
\# test       &   9,824 &   2,325 &    722 &  42,970 &    69 &   8,880 \\
\hline
\end{tabular}
\end{center}
\caption{The number of examples in the original SNLI and FEVER data, and the
number of examples generated for each adversarial distraction.}
\end{table*}

\subsection{Passive voice}

We first evaluate and teach the use of the passive voice.
By changing a hypothesis to passive, we can obtain a semantically
equivalent hypothesis with identical tokens except for a different
conjugation of the verb, the insertion of the word ``by,'' and a
different word order.

To perform the conversion, we use semantic role labeling results
of SENNA \citep{senna} to identify verbs and their ARG0 and ARG1
relations.  We change the form of the verb and move the ARG0 and ARG1
phrases to opposite sides of the verb to form the passive.
Here, we use head words
identified in dependency parsing results and the part-of-speech tagging 
information of the verb from spaCy \citep{spacy}
to change the verb form correctly according to the plurality of these
nouns and the tense of the verb. The transformation is applied only 
at the root verb identified in dependency parsing output.

If the addition of the passive were the only augmentation, models would
not be learning that word order matters.  Thus, in the cases where
the original pair is an entailment, we add an adversarial
distraction where the label is contradiction, by reversing the subject
and object in the hypothesis after transformation to passive.
We call this the {\em passive reversal}.
We filter out cases where the root verb in the hypothesis is a reciprocal
verb, such as ``meet'' or ``kiss,'' or a verb appearing with the preposition
``with,'' so that
the resulting sentence is surely not implied by the premise if the original
is. 
For example, the hypothesis, ``A woman is using a large umbrella'' (entailment),
generates the passive example, ``A large umbrella is being used by a woman''
(entailment), and the passive reversal, ``A woman is being used by a large
umbrella'' (contradiction).

\subsection{Person reversal}

One weakness of adversarial distraction by passive reversal is that
many hypotheses become ridiculous.  A model leveraging language model
information can guess that a hypothesis such as ``A man is being worn
by a hat'' is a contradiction without considering the premise.  Indeed,
when we train a hypothesis only baseline \citep{hypothesisOnly18} with
default parameters using the SNLI dataset augmented with passive and
passive reversal examples, 95.98\% of passive reversals are classified correctly
from the hypothesis alone, while only 67.89\% of the original and 64.69\% 
of the passive examples are guessed correctly.

To generate more plausible adversarial distractions, we try reversing
named entities referring to people.  Most person names should be
equally likely with or without reversal, with respect to a language model,
so the generated examples should rely on understanding the syntax of the
premise correctly.

SNLI generally lacks named entities, because it is
sourced from image captions, so we consider FEVER data instead.
The baseline FEVER system \citep{fever18} retrieves up to five sentences
of potential evidence from Wikipedia for each claim, by comparing TFIDF
scores.  We label each of these
evidence/claim pairs as entailment or contradiction, according to the claim
label, if the evidence appears in the ground truth evidence set,
and as neutral otherwise.  Because the potential evidence is pulled
from the middle of an article, it may be taken out of context with
coreference relations unresolved.  To provide a bit of context,
we prefix each premise with the title of the Wikipedia page it
comes from, punctuated by brackets.

Our {\em person reversal} dataset generates contradictions from entailment
pairs, by considering person named entities identified by SENNA in the
hypothesis, and reversing them if they appear within the ARG0 and ARG1 phrases
of the root verb.  Again, we filter examples with reciprocal verbs
and the preposition ``with.'' For example, the FEVER claim, ``Lois Lane's name was taken from Lola
Lane's name'' (entailment), leads to a person reversal of 
``Lola Lane's name was taken from Lois Lane's name'' (contradiction).

To compare the plausibility of the examples, we train the same hypothesis
only baseline on the FEVER dataset augmented with the person reversals.
It achieves 15.94\% accuracy on the added examples, showing that
the person reversals are more plausible than the passive reversals.

\subsection{Life spans}

Our third adversarial distraction ({\em birthday}) involves distinguishing
birth and death date information.  It randomly inserts birth and death dates
into a premise
following a person named entity, in parentheses, using one of two date formats.
If it chooses a future death date, no death date is inserted.
A newly generated hypothesis randomly gives a statement about either birth or
death, and either the year or the month, with a label equally balanced among
entailment, contradiction, and neutral.  For half of the contradictions it
simply reverses the birth and death dates; otherwise it randomly chooses
another date.  For the neutral examples it asks about a different named entity,
taken from the same sentence if possible.

For example, a birthday and death date are randomly generated to yield
the premise,
``[Daenerys Targaryen] Daenerys Targaryen is a fictional character in George R. R. Martin's A Song of Ice and Fire series of novels, as well as the television adaptation, Game of Thrones, where she is portrayed by Emilia Clarke (April 25, 860 -- November 9, 920),'' and hypothesis, ``Emilia
Clarke died in April'' (contradiction).

\section{Evaluation}

\begin{table*}[t!]
\begin{center}
\begin{tabular}{lcccccc}
\hline
 & \multicolumn{3}{c}{SNLI} & \multicolumn{3}{c}{FEVER} \\
 & Original & Passive & Passive rev. & Original & Person rev. & Birthday \\
\hline \hline
DA   & .8456 & .8301 & .0111 & .8416 (.1503) & .0435 & .2909 \\
ESIM & .8786 & .8077 & .0139 & .8445 (.3905) & .0290 & .3134 \\
FTLM & .8980 & .8430 & .0540 & .9585 (.6656) & .0000 & .2953 \\
\hline
\end{tabular}
\end{center}
\caption{Accuracy and (Cohen's Kappa) when training on original SNLI or FEVER data and testing on original or added examples.}
\end{table*}

\begin{table*}[t!]
\begin{center}
\begin{tabular}{lccccccc}
\hline
 & \multicolumn{3}{c}{SNLI} & \multicolumn{2}{c}{Person rev.} & \multicolumn{2}{c}{Birthday}\\
 & Original & Passive & Passive rev. & Original & Added & Original & Added \\
\hline \hline
DA   & .8517 & .7333 & .5042 & .8552 (.1478) & .1449 & .8925 (.1550) & .4700 \\
ESIM & .8781 & .8667 & .9833 & .8406 (.3809) & .6232 & .8721 (.4404) & .9684 \\
FTLM & .8953 & .8920 & .9917 & .9581 (.6610) & .7536 & .9605 (.6809) & .9926 \\
\hline
\end{tabular}
\end{center}
\caption{Accuracy and (Cohen's Kappa) when training on augmented SNLI (SNLI + passive + passive reversal) or augmented FEVER (FEVER + person reversal or FEVER + birthday) and testing on original or added examples.}
\end{table*}

\subsection{Experiments}

We consider three NLI systems based on deep neural networks:
Decomposable Attention (DA) \citep{DecAtt16}, ESIM \citep{esim17},
and a Finetuned Transformer Language Model (FTLM) \citep{openai_trans18}.
For Decomposable Attention we take the AllenNLP implementation
\citep{gardner17} without ELMo features \citep{peters18};
for the others, we take the releases from the authors.
We modify the code released for FTLM to support entailment
classification, following the description in the paper.

For the FEVER-based datasets, for DA and ESIM, we reweight each class
in the training data in inverse proportion to its number of examples.
This reweighting is necessary to produce nontrivial (most frequent class)
results; the NLI training set we derive from FEVER has 92\% neutral,
6\% entailment, and 2\% contradiction examples.  FTLM requires no such
reweighting.  When evaluating on the original FEVER examples, we report
Cohen's Kappa between predicted and
ground truth classifications, in addition to accuracy, because the imbalance
pushes DA and ESIM below the accuracy of a trivial classifier.

Whereas FTLM uses a byte-pair encoding vocabulary \citep{bpe16} that can
represent any word as a combination of subword tokens, DA and ESIM rely
on word embeddings from GloVe \citep{glove14}, with a single
out-of-vocabulary (OOV) token shared for all unknown words.  Therefore it is
unreasonable to expect DA and ESIM not to confuse named entities in
FEVER tasks.  We extend each of these models by allocating 10,000 random
vectors
for out-of-vocabulary words, and taking a hash of each OOV word to select one of
the vectors.  The vectors are
initialized from a normal distribution with mean 0 and standard deviation 1.

\subsection{Results}

When we train the three models using original SNLI without augmentation,
the models have slightly lower performance on the passive examples than
the original data.  However, all three models fail to properly classify
the passive reversal data: without training, it looks too similar to the
original hypothesis.
The augmented data succeeds in training two out of three of the
models about the passive voice: ESIM and FTLM can classify the passive
examples with approximately the accuracy of the original examples, and
with even higher accuracy, they can pick out the passive reversals
as preposterous.  However, DA cannot do better than guess whether
a passive sentence is reversed or not.  This is because its model
is defined so that its output is invariant to changes in word order.  Because
it must consider the possibility of a passive reversal, its performance
on the passive examples actually goes down after training with augmentations.

Person reversal also stumps all three models before retraining.
Of course DA can get a person reversal right only when it gets an
original example wrong, because of its insensitivity to word order.
ESIM and FTLM find person reversals to be more difficult
than the original examples.  Compared to passive reversal, the lack of
language hints seems to make the problem more challenging.  However, the
multitude of conditions necessary to perform a person reversal
makes the added examples less than 1\% of the overall training data.

No model trained on FEVER initially does better than random guessing on the
birthday problem.  By training with augmented examples, ESIM and FTLM learn
to use the structure of the premise properly to solve this problem. DA learns
some hints after retraining, but essentially the problem depends on word order,
which it is blind to.  It is noteworthy that the performance of all three
models on the original data improves after the birthday examples are added,
unlike the other two augmentations, where performance remains the same.
Four percent of the original FEVER claims use the word ``born'' or ``died,''
and extra practice with these concepts proves beneficial.

\section{Discussion}

We have taken two basic aspects of syntax, the equivalence of passive
and active forms of the same sentence, and the distinction between
subject and direct object, and shown that they are not naturally learned
through existing NLI training sets.  Two of the models we evaluated
could master these concepts with added training data, but the
Decomposable Attention model could not even after retraining.

We automatically generated training data to teach these syntactic concepts,
but doing so required rather complicated programs to manipulate sentences
based on parsing and SRL results.  Generating large numbers of
examples for more complicated or rarer aspects of syntax will be
challenging.  An obvious difficulty in extending our approach
is the need to make a distraction template that affects the meaning
in a known way.  The other difficulty lies in making transformed
examples plausible enough not to be rejected by language model likelihood.

\bibliography{augmentation}
\bibliographystyle{aclnatbibnourl}

\end{document}